\documentclass[twoside]{arxiv}

\usepackage[T1]{fontenc}
\usepackage[english]{babel}
\usepackage[utf8]{inputenc}
\usepackage{amsmath,amssymb,amsthm,dsfont}
\usepackage{algorithm}
\usepackage[noend]{algpseudocode}
\usepackage[colorinlistoftodos]{todonotes}
\usepackage{booktabs}
\usepackage{url}
\usepackage{accents}
\usepackage{subcaption}
\usepackage{graphicx}

\usepackage{amsmath,amsfonts,bm}

\def\eqref#1{equation~\ref{#1}}
\def\1{\bm{1}}

\DeclareMathAlphabet{\mathsfit}{\encodingdefault}{\sfdefault}{m}{sl}
\SetMathAlphabet{\mathsfit}{bold}{\encodingdefault}{\sfdefault}{bx}{n}

\DeclareMathOperator*{\argmin}{arg\,min}

\newcommand*\Let[2]{\State #1 $\gets$ #2}
\newtheorem{remark}{Remark}

\newlength{\dhatheight}
\newcommand{\doublehat}[1]{%
    \settoheight{\dhatheight}{\ensuremath{\hat{#1}}}%
    \addtolength{\dhatheight}{-0.35ex}%
    \hat{\vphantom{\rule{1pt}{\dhatheight}}%
    \smash{\hat{#1}}}}

\def\lgrad{{\sc lossgrad}}
\newcommand\il[1]{\langle #1 \rangle}
\def\wngrad{{\sc wngrad }}
\def\l4{{\sc l4 }}

\begin{document}

\author{Bartosz Wójcik$^1$, Łukasz Maziarka$^2$, Jacek Tabor$^2$\\{\small\rm $^1$Faculty of Physics, Mathematics and Computer Science \\ Cracow University of Technology \\ $^2$Faculty of Mathematics and Computer Science \\ Jagiellonian University \\ e-mail: {\it bartwojc@gmail.com, l.maziarka@gmail.com, jcktbr@gmail.com}}}

\title{{\large\bf \lgrad{}: automatic learning rate in gradient descent}}

\maketitle

\abstract{In this paper, we propose a simple, fast and easy to implement algorithm \lgrad{} (locally optimal step-size in gradient descent), which automatically modifies the step-size in gradient descent during neural networks training. Given a function $f$, a point $x$, and the gradient $\nabla_x f$ of $f$, we aim to find the step-size $h$ which is (locally) optimal, i.e. satisfies:
$$
h=\argmin_{t \geq 0} f(x-t \nabla_x f).
$$
Making use of quadratic approximation, we show that the algorithm satisfies the above assumption.
We experimentally show that our method is insensitive to the choice of initial learning rate while achieving results comparable to other methods.
}


\section{Introduction}

Gradient methods, with the stochastic gradient descent at the head, play a basic and crucial role in nonlinear optimization of artificial neural networks.
In recent years many efforts have been devoted to better understand and improve existing optimization methods. This led to the widespread use of the Momentum method \cite{qian1999momentum} and learning rate schedulers \cite{xing2018walk}, as well as to creation of new algorithms, such as AdaGrad \cite{duchi2011adaptive}, AdaDelta \cite{zeiler2012adadelta}, RMSprop \cite{tieleman2012lecture}, Adam \cite{kingma2014adam}.

These methods, however, are not without flaws, as they need an extensive tuning or costly grid search of hyperparameters, together with suitable learning rate scheduler -- too large step-size may result in instability, while too small may be slow in convergence and may lead to stuck in the saddle points \cite{ruder2016overview}.

As the choice of the proper learning rate is crucial, in this paper we aim to find such step which is locally optimal with respect to the direction of the gradient. Our ideas were motivated by \l4 algorithm \cite{rolinek2018l4}, which however apply the idea globally, as it computes the linearization of the loss function at
the given point and proceeds to the global root of this linearization. Furthermore, unlike the \wngrad \cite{wu2018wngrad}, our algorithm can both increase and decrease the learning rate values.

Our algorithm is similar in principle to the line search methods as its aim is to adjust the learning rate based on function evaluations in the selected direction. Line search methods however, perform well only in a deterministic setting and require a stochastic equivalent to perform well with SGD \cite{mahsereci2015probabilistic}. In opposite to this, \lgrad{} is also intended to work well in a stochastic setting. The difference is that our algorithm changes the step-size after taking the step thus never requiring any copying of potentially large amounts of memory for network weights.

\section{LOSSGRAD}

The method we propose is based on the idea of
finding a step-size which is locally optimal, i.e. we follow the direction of the gradient to maximally minimize the cost function. Thus given a function $f$ (which we want to minimize), a point $x$ and the gradient\footnote{Clearly, we can use an arbitrary direction provided by some minimization method instead of the gradient in place of $v$} $\nabla_x f$, we aim to find
the step-size $h$ which is (locally) optimal, i.e. satisfies:
\begin{equation} \label{eq:1}
h_{opt} = \argmin_{t \geq 0} f(x-t v), \text{ where }
v=\nabla_x f.
\end{equation}
A natural problem of how to practically realize the above search emerges. This paper is devoted to the examination of one of the possible solutions.

\begin{figure}
    \centering
    \includegraphics[width=.5\linewidth]{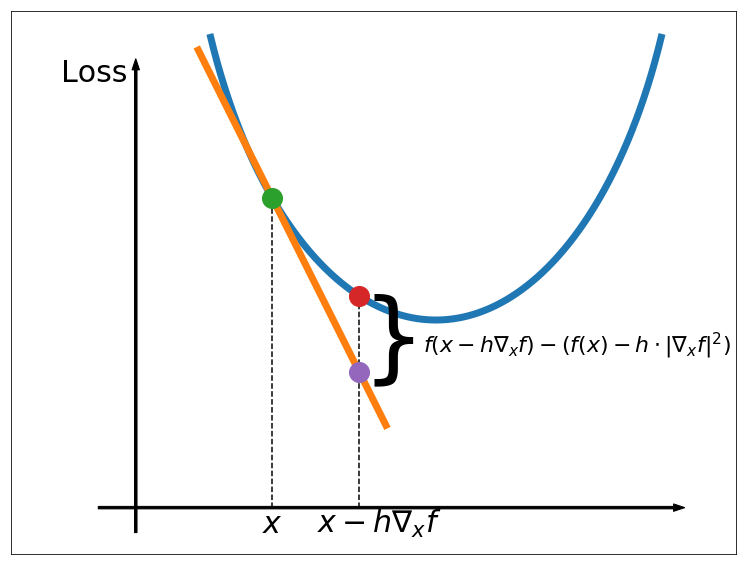}
    \caption{$r_h$ measures the relation between the true value of the loss function $f$ in point $x - h \il{v,\nabla _x f}$ and its linearized prediction given by the gradient.}
    \label{fig:r_h}
\end{figure}

We assume that we are given a candidate $h>0$ from the previous step (or some initial choice in the first step). 
A crucial role is played by investigation of the connection between the value of $f$ after the step size and the value given by its linearized prediction (see Figure \ref{fig:r_h}):
$$
r_h=\frac{f(x-hv) - (f(x) - h \il{v,\nabla _x f})}{h \il{v,\nabla _x f}}.
$$
We implicitly assume here that $\il{\nabla f_x,v} >0$ (if this is not the case, we replace $v$ by $-v$).

\begin{figure}
    \centering
    \includegraphics[width=.5\linewidth]{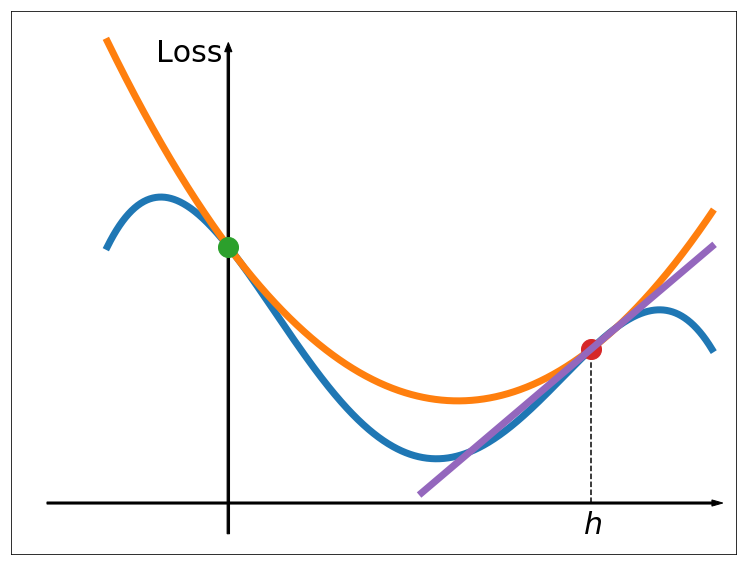}
    \caption{Function $\phi$ with its corresponding quadratic approximation $W$. If the derivative of $W$ at point $h$ is positive, we should decrease the step-size to obtain the minima of $W$.}
    \label{fig:W_phi}
\end{figure}

Our idea relies on considering the loss function in a direction $v$:
$$
\phi:t \to f(x-tv),
$$
and fitting a quadratic function $W(t)$, which coincides with $\phi$ at the end points and has the same derivative at zero (see Figure \ref{fig:W_phi}), i.e. such that:
$$
\phi(0)=W(0), \, \phi'(0)=W'(0), \, \phi(h)=W(h).
$$
Then we get:
\begin{equation} \label{eq:2}
\phi(t) \approx W(t) = f(x)-t \il{\nabla_x f,v} +r_h \frac{\il{\nabla_x f, v}}{h} t^2.
\end{equation}

\begin{remark}
To compute $r_h$ we need the knowledge of the gradient $\nabla_x f$ and the evaluation of $f$ at $x-hv$ (the predicted next point in which we will arrive according to the current value of the step-size). Consequently, in the case when $v=\nabla_x f$ (i.e. in the case of gradient methods), we need to additionally compute $f(x-h\nabla_x f)$. Then the value $r_h$ simplifies to:
$$
r_h=\frac{f(x-h\nabla_x f) - (f(x) - h \cdot \|\nabla_x f\|^2)}{h \cdot \|\nabla_x f\|^2}.
$$
\end{remark}

\noindent The derivative of function $W$ at point $h$ is given by:
\begin{equation} \label{eq:D}
W'(h) = - \il{\nabla_x f,v}+r_h \frac{\il{\nabla_x f,v}}{h} 2 h = \il{\nabla_x f,v} \cdot (-1+2r_h).
\end{equation}

\noindent This means that if:
$$
-1+2r_h \leq 0 \text{ i.e. } r_h \leq 1/2,
$$
the derivative of function $W$ at point $h$ is negative, and therefore we should increase the step-size to further minimize $f$ (see Figure \ref{fig:W_phi}). On the other hand, if :
$$
-1+2r_h > 0 \text{ i.e. } r_h > 1/2,
$$
the derivative of function $W$ at point $h$ is positive, and therefore we should decrease the step-size to further minimize $f$.

In our method, increasing (decreasing) the step-size takes place by multiplication of the current learning rate $h$ by the learning rate adjustment factor $c$ ($\frac{1}{c}$). One can find the \lgrad{} algorithm pseudocode in algorithm \ref{alg:lossgrad-alg}. Notice that using our method does not involve almost any additional calculations.

\begin{algorithm}
  \caption{\lgrad{} step}
  \label{alg:lossgrad-alg}
  \begin{algorithmic}[1]
    \Require{$X$ - inputs for current batch, $y$ - labels for current batch}
    \Require{$\theta$ - model weights, $\alpha$ - learning rate, $c$ - learning rate adjustment factor}
    \Statex
    \Function{lossgrad\_step}{$X, y$}
      \Let{$\hat{y}$}{$\text{predict}(X; \theta)$}
      \Let{$f$}{$\text{loss\_function}(\hat{y}, y)$}
      \Let{$\text{approx}$}{$f - h ||\nabla_\theta f||^2$}
      \Let{$\theta$}{$\theta - h \nabla_\theta f$}
      \Let{$\doublehat{y}$}{$\text{predict}(X; \theta)$}
      \Let{$\text{actual}$}{$\text{loss\_function}(\doublehat{y}, y)$}
      \Let{$r_h$}{$\frac{\text{actual} - \text{approx}}{h ||\nabla_\theta f||^2}$}
      \If {$r_h > 0.5$}
        \Let{$\alpha$}{$\frac{\alpha}{c}$}
      \Else
        \Let{$\alpha$}{$c\alpha$}
      \EndIf
    \EndFunction
  \end{algorithmic}
\end{algorithm}

\section{LOSSGRAD asymptotic analysis in two dimensions}

In the following section we show, how this process behaves for the quadratic form $F(x)=x^TAx$, where $x = (x_1, \ldots, x_n) \in \mathbb{R}^n$ and $A$ is a symmetric positive matrix. Observe that in this case \lgrad{} can be seen as the approximation of the exact solution to equation (\ref{eq:1}). Therefore in this section, we study how the minimization process given in (\ref{eq:1}) works for quadratic functions.

To obtain exact formula we now apply the orthonormal change of coordinates in which $F$
has the simple form:
$$
F(x_1,x_2,\ldots)=\lambda_1 x_1^2+\lambda_2 x_2^2+\ldots + \lambda_n x_n^2,
$$
where $\lambda_1 \geq \lambda_2 \geq \ldots \lambda_n \geq 0$ are the eigenvalues of $A$.

Starting from the random point $x^0=(x^0_1, x^0_2, \ldots, x^0_n)$, which gradient is equal to $ \nabla^0=2(\lambda_1 x^0_1,\lambda_2 x^0_2,\ldots)$, we have:
$$
f(t) = F(x^0-t \nabla^0)=F((1-2t\lambda_1)x_1^0, (1-2t\lambda_2)x_2^0, \ldots, (1-2t\lambda_n)x_n^0) =
$$
$$
= \lambda_1[(1-2t\lambda_1)x_1^0]^2 + \lambda_2[(1-2t\lambda_2)x_2^0]^2 + \ldots + \lambda_n[(1-2t\lambda_2)x_n^0]^2.
$$

\noindent After taking the derivative we have:
$$
f'(t) = -4\lambda_1^2(1-2t\lambda_1)(x_1^0)^2 - 4\lambda_2^2(1-2t\lambda_2)(x_2^0)^2 - \ldots - 4\lambda_n^2(1-2t\lambda_n)(x_n^0)^2 
$$
$$
= -4[\lambda_1^2(x_1^0)^2 + \lambda_2^2(x_2^0)^2 + \ldots + \lambda_n^2(x_n^0)^2] + 8t[\lambda_1^3(x_1^0)^2 + \lambda_2^3(x_2^0)^2 + \ldots + \lambda_n^3(x_n^0)^2].
$$

\noindent As our goal is to find $t_0 = \argmin_{t} f(t)$, we equate the derivative to zero:
$$
t_0 = \frac{1}{2} \frac{\lambda_1^2(x_1^0)^2 + \lambda_2^2(x_2^0)^2 + \ldots + \lambda_n^2(x_n^0)^2}{\lambda_1^3(x_1^0)^2 + \lambda_2^3(x_2^0)^2 + \ldots + \lambda_n^3(x_n^0)^2}.
$$
Since $x^1=x^0-t_0\nabla^0$, we have:
$$
x^1 = \left(\frac{(\lambda_1-\lambda_1)\lambda_1^2(x_1^0)^2+ (\lambda_2-\lambda_1)\lambda_2^2(x_2^0)^2+\ldots}{\lambda_1^3(x_1^0)^2+
\lambda_2^3(x_2^0)^2+\ldots}\cdot x_1^0,\ldots \right)
$$

To see how this process behaves after a greater number of steps, we assume that $x \in \mathbb{R}^2$. We consider the function which transports the point to its next iteration:
$$
g(x)=\frac{\lambda_2-\lambda_1}{\lambda_1^3(x_1)^2+
\lambda_2^3(x_2)^2}x_1x_2(\lambda_2^2 x_2,-\lambda_1^2 x_1).
$$
Then
$$
g^2(x)=\frac{\lambda_2-\lambda_1}{\lambda_1^3(x_1)^2+
\lambda_2^3(x_2)^2}x_1x_2 \cdot g(\lambda_2^2 x_2,-\lambda_1^2 x_1)
$$
$$
=\frac{(\lambda_2-\lambda_1)}{\lambda_1^3(x_1)^2+
\lambda_2^3(x_2)^2}x_1x_2
\frac{(\lambda_2-\lambda_1)}{\lambda_1^3(\lambda_2^2 x_2)^2+
\lambda_2^3(\lambda_1^2 x_1)^2}\lambda_1^2\lambda_2^2 x_1x_2
\lambda_1^2\lambda_2^2 (x_1,x_2)
$$
$$
=\frac{(\lambda_2-\lambda_1)^2\lambda_1\lambda_2}{\lambda_1^3(x_1)^2+
\lambda_2^3(x_2)^2}
\frac{x_1^2x_2^2}{\lambda_1 x_1^2+\lambda_2 x_2^2} 
 (x_1,x_2)=\frac{(\lambda_2-\lambda_1)^2\lambda_1\lambda_2}{(\lambda_1^2+\lambda_2^2)\lambda_1\lambda_2+\lambda_1^4 a_x+\lambda_2^4 a_x^{-1}} (x_1,x_2),
$$
where $a_x=x_1^2/x_2^2$, and thus $g^{2n}(x)=K^n_x x$. By taking the worst possible case we obtain that the above minimizes for $a_x = \lambda_2^2/\lambda_1^2$, so we obtain an estimate for the convergence:
\begin{equation} \label{e:oszac}
\|g^{2n}(x)\| \leq \big(\tfrac{\lambda_1-\lambda_2}{\lambda_1+\lambda_2}\big)^{2n}\|x\|.
\end{equation}
Notice that this is invariant to data and function scaling (i.e. $g(Cx)=Cg(x)$).

\begin{remark}
One can easily observe that the estimation (\ref{e:oszac}) gives the upper bound for a decrease rate of the solution to any standard gradient descent method with a fixed learning rate. Whether the same holds for the method in $\mathbb{R}^n$ is an open problem.
\end{remark}

\section{Experiments}

We tested \lgrad{} in multiple different setups. First, we explore the algorithm's resilience on the choice of initial learning rate and its behavior for a range of different LR adjustment factor values. Then we proceed to test the algorithm's performance on fully connected networks on MNIST classification and Fashion-MNIST \cite{xiao2017fashion} autoencoder tasks. Convolutional neural networks are tested on CIFAR-10 \cite{krizhevsky2009learning} classification task using ResNet-56 architecture \cite{he2016deep}. We also evaluate our optimizer for an LSTM model \cite{hochreiter1997long} trained on Large Movie Review Dataset (IMDb) \cite{maas2011learning}. Finally, we test \lgrad{} on Wasserstein Auto-Encoder with Maximum Mean Discrepancy-based penalty \cite{tolstikhin2017wasserstein} on CelebA dataset \cite{liu2015deep}. Network architectures and all hyperparameters used in experiments are listed in the appendix \ref{sec:architecture-and-hyperparameters}

\begin{figure}[h]
	\centering
	\includegraphics[width=0.495\linewidth,trim={3.0cm 0.25cm 4.25cm 1.25cm},clip]{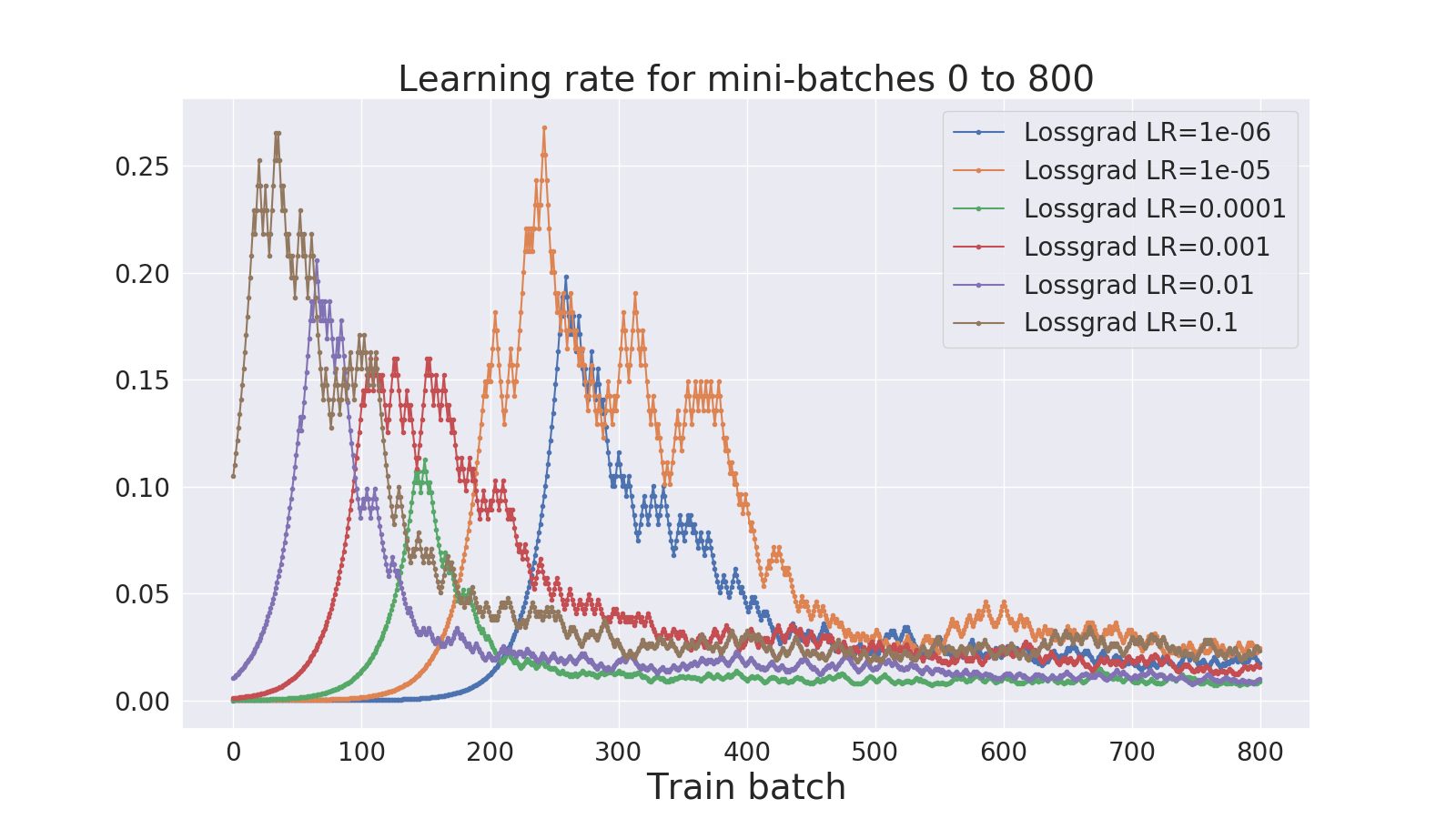}
	\includegraphics[width=0.495\linewidth,trim={3.0cm 0.25cm 4.25cm 1.25cm},clip]{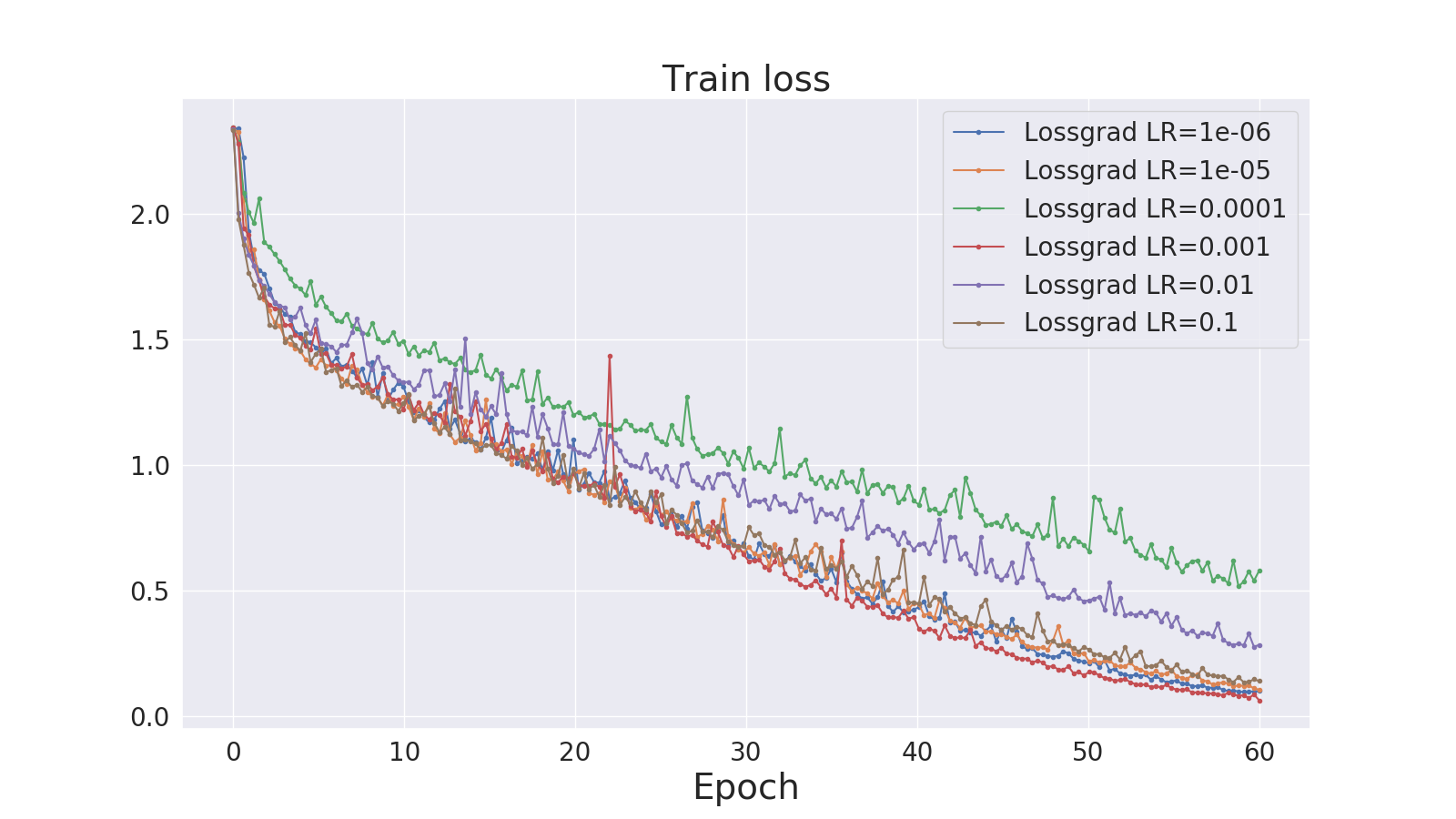}
	\caption{\lgrad{} with different initial learning rate values trained on CIFAR-10 dataset.}
	\label{fig:cifar-convergence}
\end{figure}

In our experiments we tried to compare \lgrad{} with \wngrad \cite{wu2018wngrad} and \l4 applied to SGD \cite{rolinek2018l4}. 
We found out that \l4 based on vanilla SGD is extremely unstable both on standard and tuned parameters on almost all datasets and network architectures, so we do not present the results here. 
For each experiment we also tested standard SGD optimizer with a range of learning rate hyperparameters, including a highly tuned one. For comparison, we also included SGD with scheduled learning rate if that enhanced the results. Because dropout heavily affects \lgrad{}'s stability, we decided not to use it in our experiments.

\begin{table}[h]
\footnotesize
\parbox{.49\linewidth}{
\centering
\begin{tabular}{| l | c | c |}
\toprule
  & Test loss & Test acc. \\
\midrule
Lossgrad c=1.001 & 2.737 & 0.655 \\
Lossgrad c=1.005 & 2.870 & \textbf{0.673} \\
Lossgrad c=1.01 & 1.995 & 0.658 \\
Lossgrad c=1.05 & \textbf{1.165} & 0.665 \\
Lossgrad c=1.1 & 1.862 & 0.661 \\
Lossgrad c=1.2 & 2.830 & 0.643 \\
\bottomrule
\end{tabular}
\caption{\lgrad{} results for different c hyperparameter trained on CIFAR-10 dataset.}
\label{tab:cifar-different-cs}
}
\hfill
\parbox{.49\linewidth}{
\centering
\begin{tabular}{| l | c | c |}
\toprule
 & Test loss & Test acc. \\
\midrule
SGD LR=0.01 & 0.101 & 0.971 \\
SGD LR=0.07 & 0.085 & 0.980 \\
SGD LR=0.45 & 0.087 & \textbf{0.985} \\
WNGrad & \textbf{0.073} & 0.978 \\
Lossgrad & 0.094 & 0.980 \\
\bottomrule
\end{tabular}
\caption{\lgrad{} results for classification on MNIST dataset.}
\label{tab:mnist-results}
}
\end{table}

We test the initial learning rate with values ranging between $10^{-1}$ and $10^{-6}$ for a convolutional neural network on CIFAR-10 with the rest of the settings staying the same. Resulting test loss values are presented in Fig. \ref{fig:cifar-convergence} on the right, while the first 800 batches' step size values are presented on the left. Irrespectively of the initial learning rate chosen, the step size always converges to values around $0.025$ for this experiment setup. Thus, the need for tuning is practically eliminated, and this property makes the algorithm noticeably attractive.

As \lgrad{} requires one hyperparameter, we explore which values are appropriate. This is tested by training a convolutional neural network on CIFAR-10 dataset, using our optimizer parameterized with a different value each time, with the rest of the settings staying the same. We evaluated the following hyperparameter values: $1.001, 1.005, 1.01, 1.05, 1.1, 1.2$. We found that low and high values tend to cause unstable behavior. According to these results, the rest of the experiments in this paper use $c=1.05$ and initial learning rate set to $1^{-4}$.

\begin{table}[h]
\footnotesize
\parbox{.49\linewidth}{
\centering
\begin{tabular}{| l | c | c |}
\toprule
  & Train loss & Test loss \\
\midrule
SGD LR=2.0 & 0.018 & 0.019 \\
SGD LR=4.0 & 0.013 & 0.013 \\
SGD LR=16.0 & \textbf{0.010} & \textbf{0.010} \\
StepLR & 0.015 & 0.016 \\
Trapezoid & 0.012 & 0.012 \\
WNGrad & 0.023 & 0.023 \\
Lossgrad & 0.022 & 0.022 \\
\bottomrule
\end{tabular}
\caption{\lgrad{} results for autoencoder on Fashion-MNIST dataset.}
\label{tab:fashionmnist-autoencoder-results}
}
\hfill
\parbox{.49\linewidth}{
\centering
\begin{tabular}{| l | c | c |}
\toprule
 & Test loss & Test accuracy \\
\midrule
SGD LR=0.001 & 0.719 & 0.762 \\
SGD LR=0.01 & 0.561 & 0.868 \\
SGD LR=0.1 & 0.384 & 0.890 \\
MultiStepLR & \textbf{0.278} & \textbf{0.934} \\
WNGrad & 0.678 & 0.870 \\
Lossgrad & 0.492 & 0.900 \\
\bottomrule
\end{tabular}
\caption{\lgrad{} results for ResNet-56 on CIFAR-10 dataset.}
\label{tab:cifar-resnet-results}
}
\end{table}

Tab. \ref{tab:mnist-results} and Tab. \ref{tab:fashionmnist-autoencoder-results} presents the results for fully connected network trained on MNIST and an autoencoder trained on Fashion-MNIST, respectively. 
We noticed the occurrence of sudden spikes in loss (classification accuracy drops and subsequent recoveries) in case of classification on MNIST, but not when training an autoencoder on Fashion-MNIST. The spikes correspond to learning rate peaks, which suggests that temporarily too high step size causes the learning process to diverge.

\begin{figure}
	\centering
	\includegraphics[width=0.495\linewidth,trim={3.0cm 0.25cm 4.25cm 1.25cm},clip]{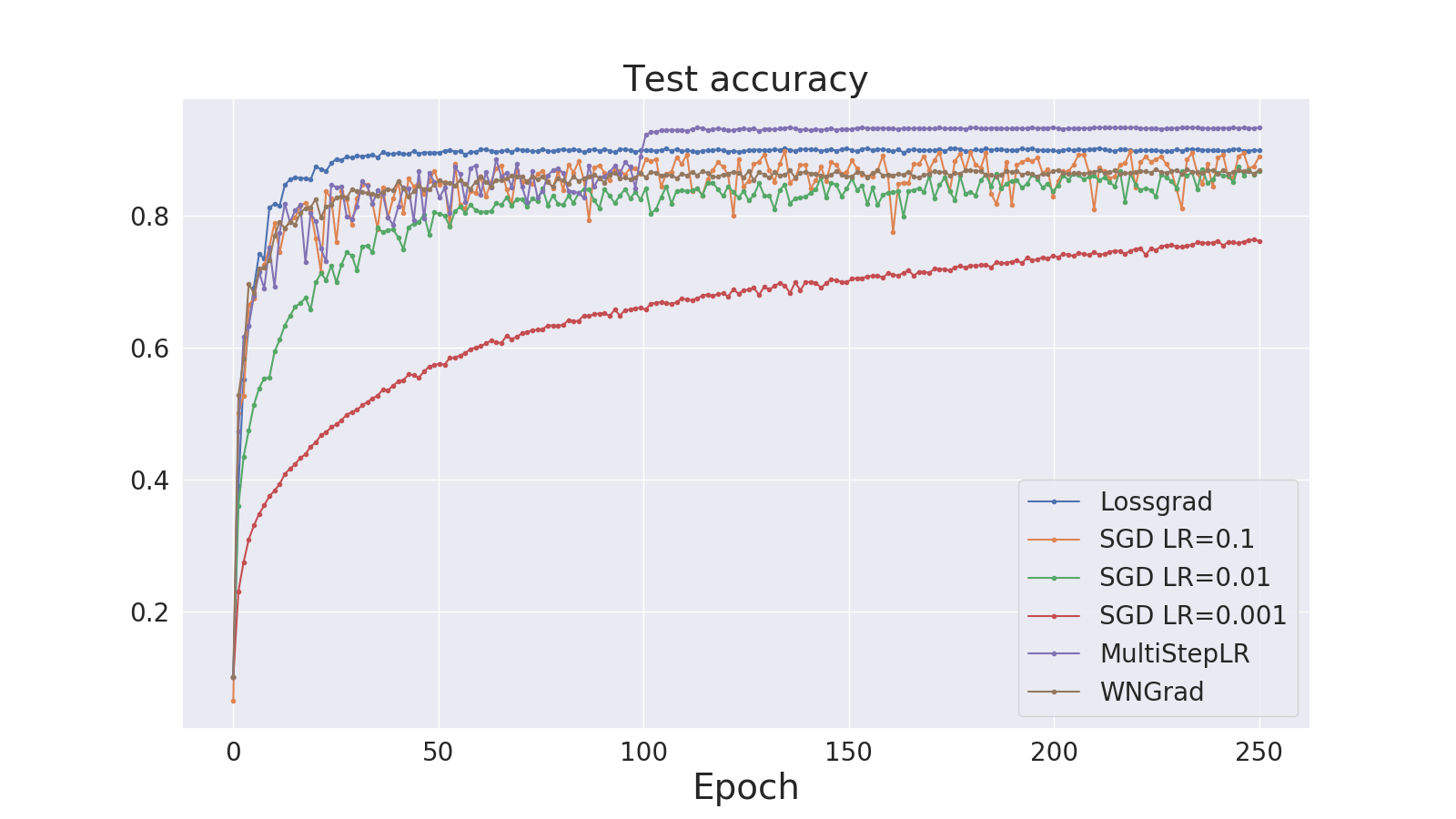}
	\includegraphics[width=0.495\linewidth,trim={3.0cm 0.25cm 4.25cm 1.25cm},clip]{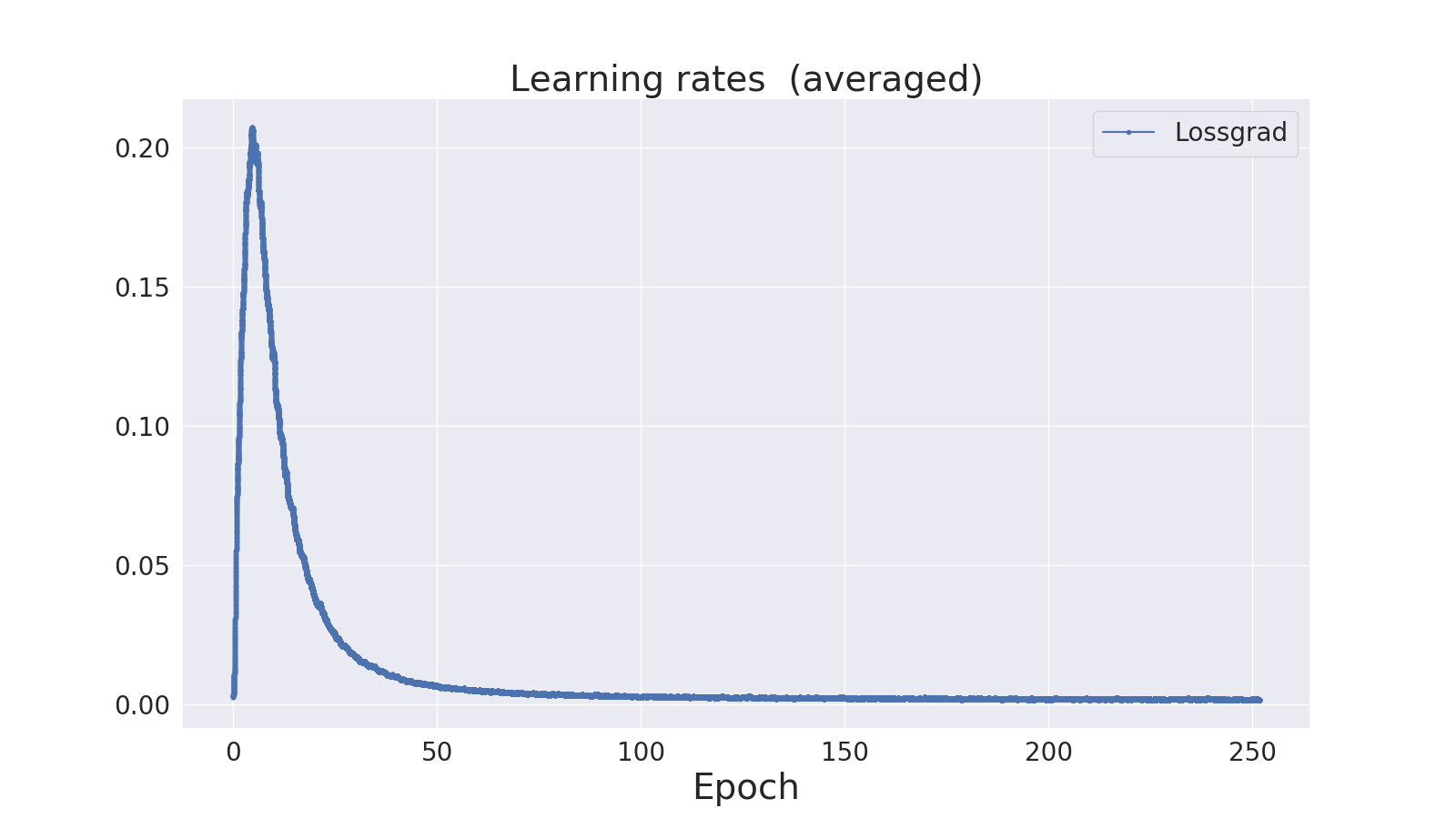}
	\caption{Training ResNet-56 with \lgrad{}.}
	\label{fig:cifar-resnet-results}
\end{figure}

Fig. \ref{fig:cifar-resnet-results} presents test accuracy and averaged step size (learning rate) when training a ResNet-56 network on CIFAR-10, while Tab. \ref{tab:cifar-resnet-results} presents the results summary. Even with low initial learning rate, \lgrad{} still manages to achieve results better than SGD, being beaten only by optimized scheduled SGD. Note the step size spike at the beginning of the training process. This spike consistently appears at the beginning of the training in nearly every setup tested in this paper. This result is in line with many learning rate schedulers used in the training of neural networks, that increase the step-size at the beginning of the training and then, after few epochs, they decrease the step-size value \cite{xing2018walk}.

\begin{table}[h]
\footnotesize
\parbox{.49\linewidth}{
\centering
\begin{tabular}{| l | c | c |}
\toprule
  & Test loss & Test acc. \\
\midrule
SGD LR=0.05 & 0.66 & 0.623 \\
SGD LR=0.1 & 0.359 & 0.845 \\
SGD LR=0.5 & \textbf{0.297} & \textbf{0.875} \\
scheduled & 0.299 & 0.874 \\
WNGrad & 0.567 & 0.726 \\
Lossgrad & 0.583 & 0.708 \\
\bottomrule
\end{tabular}
\caption{\lgrad{} result on IMDb dataset.}
\label{tab:imdb-results}
}
\hfill
\parbox{.49\linewidth}{
\centering
\begin{tabular}{| l | c | c | c | c |}
\toprule
  & Train loss & Test loss \\
\midrule
SGD LR=0.0001 & 11742.473 & 12859.591 \\
SGD LR=1e-05 & 12704.917 & 12881.991 \\
original (Adam) & \textbf{8598.712} & \textbf{11082.079} \\
WNGrad & 14321.215 & 14304.257 \\
Lossgrad & 25225.673 & 25196.921 \\
\bottomrule
\end{tabular}
\caption{\lgrad{} results for WAE on CelebA dataset.}
\label{tab:celeba-results}
}
\end{table}

Results for LSTM trained on IMDb dataset are presented in Tab. \ref{tab:imdb-results}. Here, for the vanilla SGD, a higher learning rate is preferred. 
\lgrad{} selects a very low step-size instead (below $0.01$ after the initial peak) and manages only to achieve better results than untuned SGD.

Finally, the results for WAE-MMD are presented in Tab. \ref{tab:celeba-results}. The originally used optimizer (Adam) and scheduler combination from \cite{tolstikhin2017wasserstein} is marked as ,,original''. Properly tuned SGD, as well as WNGrad, yield better results than \lgrad{}, which chooses a very low step size for this problem.

We provide an implementation of the algorithm with basic examples of usage on a git repository: \url{https://github.com/bartwojcik/lossgrad}.

\section{Conclusion}

We proposed \lgrad{}, a simple optimization method for approximating locally optimal step-size. We analyzed the algorithm behavior in two dimensions quadratic form example and tested it on a broad range of experiments. Resilience on the choice of initial learning rate and the lack of additional hyperparameters are the most attractive properties of our algorithm.

In future work, we aim to investigate and possibly mitigate the loss spikes encountered in the experiments, as well as work on increasing the algorithm's effectiveness. A version for momentum SGD and Adam is also an interesting topic for exploration that we intend to pursue.

\bibliographystyle{plain} 
\bibliography{lossgrad_arxiv}

\appendix

\section[A]{Network architecture, hyperparameters and datasets description} \label{sec:architecture-and-hyperparameters}

\begin{table}[h]
\footnotesize
\centering
\begin{tabular}{| cc | cc | ccc |}
\toprule
\multicolumn{2}{| c |}{MNIST} & \multicolumn{2}{| c |}{Fashion-MNIST} & \multicolumn{3}{| c |}{CIFAR}\\
\midrule
Type  & Outputs & Type & Outputs & Type & Kernel & Outputs \\
\midrule
Linear & 500 & Linear & 200 & Conv2d & 5x5 & 28x28x30\\
ReLU & & ReLU & & MaxPool2d & 2x2 & 14x14x30\\
Linear & 300 & Linear & 100 & ReLU & &\\
ReLU & & ReLU & & Conv2d & 5x5 & 10x10x40\\
Linear & 100 & Linear & 50 & MaxPool2d & 2x2 & 5x5x40\\
ReLU & & ReLU & & ReLU & &\\
Linear & 10 & Linear & 100 & Conv2d & 3x3 & 3x3x50\\
 & & ReLU & & ReLU & &\\
 & & Linear & 200 & Linear &  & 250\\
 & & ReLU & & ReLU & &\\
 & & Linear & 784 & Linear &  & 100\\
 & & Sigmoid & & ReLU & &\\
 & & & & Linear &  & 10\\
\bottomrule
\end{tabular}
\caption{Architecture summary for experiments presented in Tab. \ref{tab:mnist-results} (left), Tab. \ref{tab:fashionmnist-autoencoder-results} (middle), Tab. \ref{tab:cifar-different-cs} and Fig. \ref{fig:cifar-convergence} (right).}
\label{tab:architectures}
\end{table}

Tab. \ref{tab:architectures} presents the architecture used for image classification on CIFAR-10 dataset. This architecture was used to obtain the results presented in Fig.  \ref{fig:cifar-convergence} and Tab. \ref{tab:cifar-different-cs}. We initialized the neuron weights using a normal distribution with a $0.05$ standard deviation and bias weights with a constant $0.2$. The minibatch size was set to 100 and cross entropy was chosen as the loss metric. The model was trained for 60 epochs with c hyperparameter set to $1.05$ in the learning rate convergence experiment and for 120 epochs with an initial learning rate of $0.1$ in the second experiment. We preprocessed the CelebA dataset by resizing its images to 64x64 and discarding labels.

Tab. \ref{tab:architectures} also contains the architectures used for image classification on MNIST dataset and for autoencoder training on Fashion-MNIST dataset. We trained both networks for 30 epochs with the rest of the hyperparameters being the same as in CIFAR-10 experiment. For MNIST, we selected the following learning rate values: $0.01, 0.07, 0.45$ for SGD and $1.0$ for WNGrad. For Fashion-MNIST, we tested the following learning rate values: $2.0, 4.0, 16.0$ for SGD and $5.0$ for WNGrad. We also evaluated two scheduled optimizers. The first one linearly increases the learning rate from $0$ to $10$ in epoch 10, stays constant until epoch 15 and decreases afterward (trapezoidal). The second one starts with $16.0$ and is multiplied by $0.5$ after 20 epochs. Mean square error was used as the loss function.

ResNet-56 architecture is described in \cite{he2016deep}. We trained the network for 250 epochs, used random cropping and inversion when training and applied weight decay with $\lambda=0.001$. The learning rate values we used were: $0.1, 0.01, 0.001$ for SGD, $0.2$ for WNGrad, $0.1$ with $0.1$ multiplier after epochs 100 and 150 for step scheduled SGD. We set the mini-batch size to 128 in this experiment. 

For IMDb experiments, we used a pretrained embedding layer, trained with GloVe algorithm \cite{pennington2014glove} on 6 billion tokens available from \verb|torchtext| library. Its 100 element output was fed to 1 layer of bidirectional LSTM with 256 hidden units for each direction. The final linear layer transformed that to scalar output. We set $0.1$ as the learning rate for \wngrad, $0.5$, $0.1$, $0.05$ for SGD and also tested a scheduled SGD with learning rate initially set to $0.5$ and then decreasing to $0.05$ after 10 epochs.

The architecture for WAE-MMD is the same as in \cite{tolstikhin2017wasserstein}. Minibatch size was set to 64, mean square error was selected as a loss function and the network was trained for 80 epochs. We set $1^{-4}$ as the initial learning rate for WNGrad and $1^{-5}, 1^{-4}$ for SGD.

\end{document}